\documentclass[final]{cvpr}

\usepackage{times}
\usepackage{epsfig}
\usepackage{graphicx}
\usepackage{amsmath}
\usepackage{amssymb}

\usepackage{times}
\usepackage{epsfig}
\usepackage{graphicx}
\usepackage{amsmath}
\usepackage{amssymb}
\usepackage{multirow}
\usepackage{xcolor}
\usepackage{graphicx}
\usepackage{amsmath}
\usepackage{amssymb}
\usepackage{wrapfig}
\usepackage{booktabs}
\usepackage{color}
\usepackage{verbatim}
\usepackage{caption}
\usepackage{ulem}
\usepackage{epsfig}
\usepackage{enumitem}
\usepackage{placeins}
\usepackage{ragged2e}
\usepackage{textcomp}
\usepackage{bm}
\usepackage{soul}

\usepackage{makecell}
\usepackage{subfigure}
\usepackage{arydshln}

\useunder{\uline}{\ul}{}
\newcommand{\unclear}[1]{\textcolor{black}{#1}}
\newcommand{\bo}[1]{\textcolor{black}{#1}}
\newcommand{\bob}[1]{\textcolor{black}{#1}}

\newcommand{\nicknameData}{SensatUrban}
\newcommand{\qy}[1]{\textcolor{black}{#1}}

\usepackage[pagebackref=true,breaklinks=true,colorlinks,bookmarks=false]{hyperref}

\def\cvprPaperID{2107} 
\def\confYear{CVPR 2021}

\begin{document}

\title{Towards Semantic Segmentation of Urban-Scale 3D Point Clouds: \\ 
A Dataset, Benchmarks and Challenges}

\author{Qingyong Hu\textsuperscript{1}, Bo Yang\textsuperscript{1,2\thanks{Corresponding author}*}, Sheikh Khalid\textsuperscript{3}, Wen Xiao\textsuperscript{4}, Niki Trigoni\textsuperscript{1}, Andrew Markham\textsuperscript{1} \\
\textsuperscript{1}University of Oxford,
\textsuperscript{2}The Hong Kong Polytechnic University,
\textsuperscript{3}Sensat Ltd, \textsuperscript{4}Newcastle University\\
{\tt\small qingyong.hu@cs.ox.ac.uk, bo.yang@polyu.edu.hk, wen.xiao@ncl.ac.uk, andrew.markham@cs.ox.ac.uk}}

\maketitle
\pagestyle{empty}
\thispagestyle{empty}

\begin{abstract}
    An essential prerequisite for unleashing the potential of supervised deep learning algorithms in the area of 3D scene understanding is the availability of large-scale and richly annotated datasets. However, publicly available datasets are either in relative small spatial scales or have limited semantic annotations due to the expensive cost of data acquisition and data annotation, which severely limits the development of fine-grained semantic understanding in the context of 3D point clouds. In this paper, we present an urban-scale photogrammetric point cloud dataset with nearly three billion richly annotated points, \qy{which is three times the number of labeled points than the existing largest photogrammetric point cloud dataset.} Our dataset consists of large areas from three UK cities, covering about 7.6 $km^2$ of the city landscape. In the dataset, each 3D point is labeled as one of 13 semantic classes. We extensively evaluate the performance of state-of-the-art algorithms on our dataset and provide a comprehensive analysis of the results. In particular, we identify several key challenges towards urban-scale point cloud understanding. The dataset is available at \url{https://github.com/QingyongHu/SensatUrban}.
\end{abstract}

\section{Introduction}
\label{sec:intro}
The three-dimensional world around us is composed of a rich variety of objects: buildings, trees, cars, and so forth, each with distinct appearance, morphology, and function. Giving machines the ability to precisely segment and label these diverse objects is of key importance to allow them to interact competently within our physical world, for applications such as object-level robotic grasping \cite{rao2010grasping}, scene-level robot navigation \cite{valada2017adapnet} and autonomous driving \cite{geiger2013vision}, or even large-scale urban 3D modeling, which is critical for the future of smart city planning and management \cite{cornelis20083d, austin2020architecting}.

\begin{figure}[t]
\centering
\includegraphics[width=0.5\textwidth]{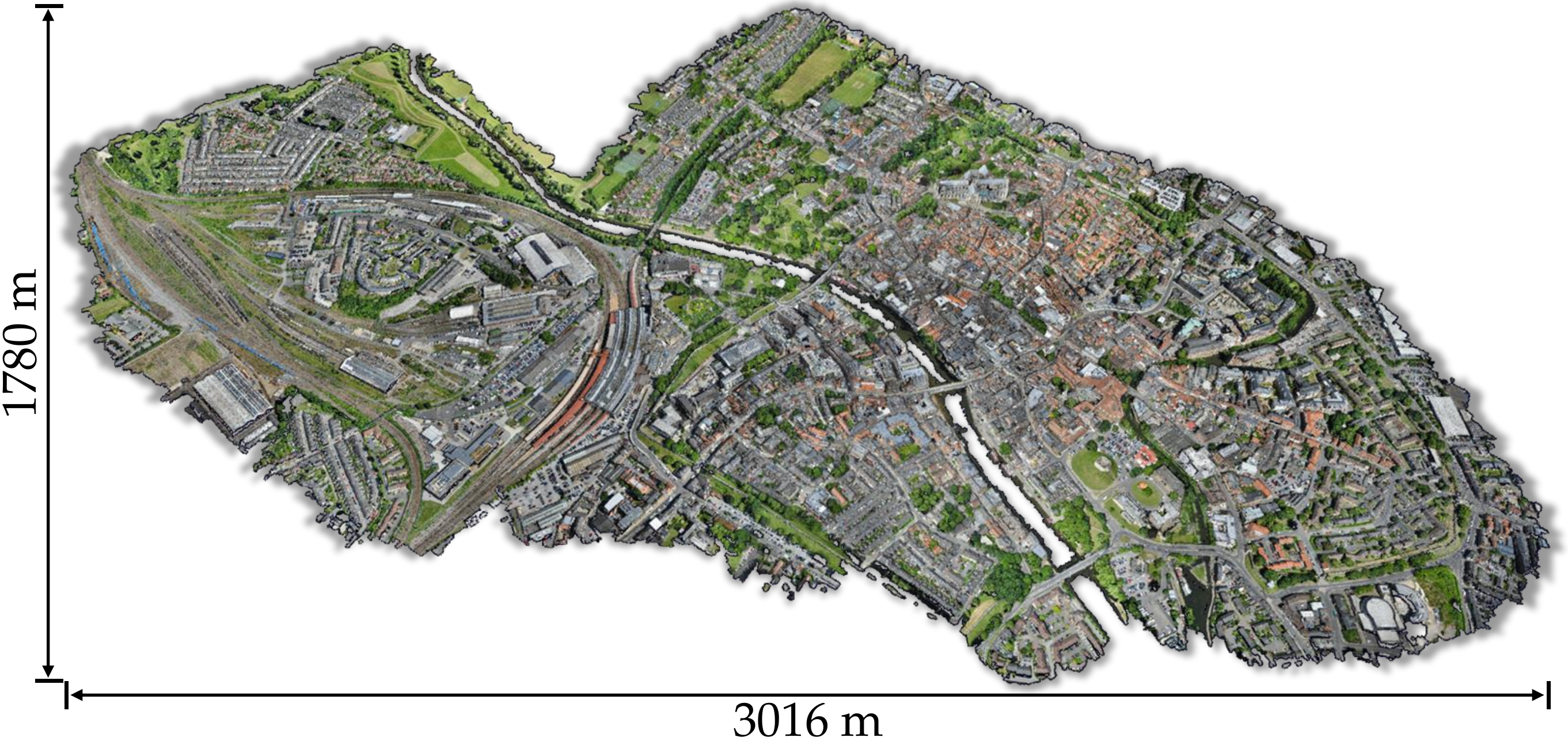}
\caption{An urban-scale point cloud collected from a region on the perimeter of the city of York, UK. It covers a contiguous area of more than 3 square kilometer and represents a typical urban suburb.}\label{fig:urban-scale}
\end{figure}

The ongoing revolution in data-driven deep networks has led to a radical boost in the performance of 3D point cloud segmentation. A series of neural pipelines proposed to address the core problem of semantic segmentation, including: 1) 3D voxel-based methods such as SparseConvNet \cite{sparse} and MinkowskiNet \cite{4dMinkpwski}, 2) 2D projection-based approaches such as RangeNet++ \cite{rangenet++} and SqueezeSeg \cite{wu2018squeezeseg}, and 3) recent point-based architectures \textit{e.g.} PointNet/PointNet++ \cite{qi2017pointnet, qi2017pointnet++}, KPConv \cite{thomas2019kpconv} and RandLA-Net \cite{hu2019randla}.

To a large degree, these techniques have been driven forward by the availability of open datasets which act as benchmarks for objective comparison of algorithms and their performance. These existing 3D repositories can be generally classified as 1) object-level 3D models such as ModelNet \cite{3d_shapenets} and ShapeNet \cite{chang2015shapenet}, 2) indoor scene-level 3D scans, \textit{e.g.,} S3DIS \cite{2D-3D-S}, ScanNet \cite{scannet}, and SceneNN \cite{scenenn}, and 3) outdoor roadway-level 3D point clouds including SemanticKITTI \cite{behley2019semantickitti} and Semantic3D \cite{Semantic3D}.


However, there remain a number of key open questions as to whether these techniques are capable of learning accurate semantics over urban-scale 3D point clouds. \textbf{Firstly}, unlike the existing datasets for objects, rooms or roadways which are usually less than 200$m$ in scale, the urban-scale datasets are expected to be collected by aerial platforms, spanning over extremely wide areas. How to efficiently and effectively preprocess massive points to feed into neural networks? \textbf{Secondly}, the real-world urban space is usually dominated by large-sized buildings or ground, and therefore the urban-scale datasets demonstrate extreme class imbalance - a majority of points fall into a few categories with sparse, yet important classes being under represented. How to overcome this data imbalance in neural networks? \textbf{Thirdly}, with the advancement of aerial mapping systems, the urban-scale point clouds can not only capture the depth information, but also true color for the scene appearance. (How) does color information, in addition to depth, aid in semantic segmentation of urban areas? \textbf{Lastly}, and potentially most importantly, how are the existing networks trained on one urban area able to generalize to a novel area?

To this end, we aim to establish a new paradigm for urban-scale 3D semantic segmentation, enabled by UAV photogrammetry. \qy{Our dataset, called \textbf{\nicknameData{}}, represents sub-sections of three large cities in the UK, \textit{i.e.}, Birmingham, Cambridge, and York. It consists of nearly four billion 3D points covering more than 7.6 square kilometers urban area in these cities (as shown in Figure \ref{fig:urban-scale}). The 3D point clouds are generated from high-quality aerial images captured by a professional-grade UAV mapping system. Details of data acquisition are presented in Section \ref{sec:dataset}. We manually labeled each point in the Birmingham and Cambridge city as one of 13 semantic categories such as \textit{ground}, \textit{vegetation}, \textit{car}, \textit{etc.}. Compared with exiting 3D datasets, our \nicknameData{} is unique in two-fold.
\begin{itemize}[leftmargin=*]
\setlength{\itemsep}{0pt}
\setlength{\parsep}{0pt}
\setlength{\parskip}{0pt}
\item Unlike existing datasets for objects \cite{3d_shapenets, chang2015shapenet}, rooms \cite{scenenn, 2D-3D-S, scannet} and roadways \cite{Semantic3D, behley2019semantickitti} which are usually less than two hundred meters in scale, the \nicknameData{} point clouds continuously occupy kilometers in real-world urban areas, opening up new opportunities towards urban-scale applications such as smart cities, and large national infrastructure planning and management. 
\item Being reconstructed from high-resolution aerial images, our point clouds provide unique top-down and oblique perspectives for the entire landscape of cities. Inherently, the geometric patterns, textures, natural colours and distributions are distinct from the existing datasets. 
\end{itemize}
}

\bo{
On the basis of \nicknameData{}, we further identify a number of key challenges and empirically investigate them from various aspects in Section \ref{sec:challenges}. In particular, we firstly study how the  large-scale urban point clouds can be pre-processed, to adapt to existing approaches without losing segmentation accuracy. Secondly, we explore the necessity of colorful appearance for better semantic learning of several key categories, highlighting the advantage of photogrammetric point clouds over LiDAR-based point clouds. Thirdly, we examine the imbalance of semantic categories in the urban-scale scenarios. Lastly, the difficulty of cross-city semantic learning is analysed. Note that, this paper does not aim to thoroughly tackle these challenges, but expose them to the community for future research.  
}

\bo{Overall, our primary contributions are: 1) a unique urban-scale 3D dataset for semantic learning, and 2) an in-depth study of generalizing existing algorithms to the large-scale urban point clouds and an outlook on future directions for 3D point cloud segmentation at massive scale and resolution. We aspire to highlight the challenges faced in the 3D semantic learning on large and dense point clouds of urban environments, sparking innovation in applications such as smart cities, digital twins, autonomous vehicles, automated asset management of large national infrastructures, and intelligent construction sites. 
}

\begin{table*}[t]
\centering
\resizebox{0.95\textwidth}{!}{%
\begin{tabular}{crcrcccc}
\Xhline{2.0\arrayrulewidth}
 & \#Name and Reference & \#Year & \#Spatial size\textsuperscript{1} & \#Classes\textsuperscript{2} & \#Points & \#RGB & \#Sensors\tabularnewline
\hline
\multirow{2}{*}{Object-level} & ShapeNet \cite{chang2015shapenet} & 2015 & - & 55 & - & No & Synthetic \tabularnewline
 & PartNet \cite{mo2019partnet} & 2019 & - & 24 & - & No & Synthetic\tabularnewline
 
\hdashline
\multirow{2}{*}{%
\begin{tabular}{c}
Indoor\tabularnewline
Scene-level\tabularnewline
\end{tabular}} & S3DIS \cite{2D-3D-S} & 2017 & 6$ \times 10^{3}m^2$ & 13 (13) & 273M & Yes & Matterport\tabularnewline
 & ScanNet \cite{scannet} & 2017 & 1.13$ \times 10^{5}m^2$ & 20 (20) & 242M & Yes & RGB-D\tabularnewline
\hdashline

\multirow{6}{*}{%
\begin{tabular}{c}
Outdoor\tabularnewline
Roadway-level\tabularnewline
\end{tabular}} & Paris-rue-Madame \cite{paris-rue-madame} & 2014 & 0.16$\times 10^{3} \ m$ & 17 & 20M & No & MLS\tabularnewline
 & IQmulus \cite{IQmulus} & 2015 & 10$\times 10^{3} \ m$ & 8 (22) & 300M & No & MLS\tabularnewline
 & Semantic3D \cite{Semantic3D} & 2017 & - & 8 (9) & 4000M & Yes & TLS\tabularnewline
 & Paris-Lille-3D \cite{NPM3D} & 2018 & 1.94$\times 10^{3} \ m$ & 9 (50) & 143M & No & MLS\tabularnewline
 & SemanticKITTI \cite{behley2019semantickitti} & 2019 & 39.2$\times 10^{3} \ m$ & 25 (28) & 4549M & No & MLS\tabularnewline
 & Toronto-3D \cite{Toronto3D} & 2020 & 1$\times 10^{3} \ m$ & 8 (9) & 78.3M & Yes & MLS\tabularnewline

\hdashline
\multirow{6}{*}{Urban-level} 
& ISPRS \cite{rottensteiner2012isprs} & 2012 & - & 9 & 1.2M & No & ALS\tabularnewline
& DublinCity \cite{zolanvari2019dublincity} & 2019 & 2$ \times 10^{6}m^2$ & 13 & 260M & No & ALS\tabularnewline
& DALES \cite{varney2020dales} & 2020 & 10 $ \times 10^{6}m^2$  & 8 (9) & 505M & No & ALS\tabularnewline
& LASDU \cite{ye2020lasdu} & 2020 & 1.02 $ \times 10^{6}m^2$  & 5 & 3.12M & No & ALS\tabularnewline
& Campus3D \cite{li2020campus3d} & 2020 & 1.58 $ \times 10^{6}m^2$ & 24 & 937.1M & Yes & UAV Photogrammetry\tabularnewline
 & \textbf{\textcolor{black}{\nicknameData{} (Ours)}} & 2020 & 7.64 $ \times 10^{6}m^2$ & 13 (31) & 2847M & Yes & UAV Photogrammetry\tabularnewline
\Xhline{2.0\arrayrulewidth}
\end{tabular}}

\caption{Comparison with the representative datasets for segmentation of 3D point clouds. \protect\textsuperscript{1}The spatial size (Area/Length) in the dataset, m: meter, \protect\textsuperscript{2} The number of classes used for evaluation and the number of sub-classes annotated in brackets. MLS: Mobile Laser Scanning system, TLS: Terrestrial Laser Scanning system, ALS: Aerial Laser Scanning system.\label{tab:Overview-dataset-1}}
\end{table*}

\section{Related Work}
\label{sec:liter}

\subsection{Existing 3D Datasets}\label{sec:liter_semseg}
\bo{Existing 3D datasets can be broadly classified into four categories: \textbf{1) Object-level 3D models}. These include the synthetic ModelNet \cite{3d_shapenets}, ShapeNet \cite{chang2015shapenet}, ShapePartNet \cite{ShapePartNet}, PartNet \cite{mo2019partnet} and the real-world ScanObjectNN \cite{scanobjectnn}. \textbf{2) Indoor scene-level 3D scans}. These datasets are usually collected by short-range depth scanners, such as NYU3D \cite{NYU3D}, SUN RGB-D \cite{sunrgbd}, S3DIS \cite{2D-3D-S}, SceneNN \cite{scenenn} and ScanNet \cite{scannet}. In addition, there are two synthetic datasets SceneNet \cite{handa2015scenenet} and SceneNet RGB-D \cite{scenenetrgbd}, which covers large-scale complex indoor environments. \textbf{3) Outdoor roadway-level 3D point clouds.} The majority of these datesets are specifically collected for applications such as autonomous driving using a LiDAR scanner together with RGB cameras, such as the early Oakland \cite{oakland}, KITTI \cite{geiger2012we}, Sydney Urban Objects \cite{de2013unsupervised} and the recent Semantic3D \cite{Semantic3D}, Paris-Lille-3D \cite{NPM3D}, Argoverse \cite{chang2019argoverse}, SemanticKITTI \cite{behley2019semantickitti}, SemanticPOSS \cite{pan2020semanticposs}, Toronto-3D \cite{Toronto3D}, nuScenes \cite{caesar2020nuscenes}, A2D2 \cite{geyer2020a2d2}, CSPC-Dataset \cite{tong2020cspc}, Lyft dataset \cite{Lyft} and Waymo dataset \cite{Waymo}. To obtain more accurate semantic labels, a number of synthetic datasets \cite{ros2016synthia, gaidon2016virtual} are generated by simulating street scenes. \qy{\textbf{4) Urban-level aerial 3D point clouds.} They are usually obtained by costly aerial LiDARs, such as the recent DublinCity \cite{zolanvari2019dublincity}, DALES \cite{varney2020dales}, LASDU \cite{ye2020lasdu}. However, they are unable to capture true color information for the complex urban structures. }}


\qy{Being concurrent to our work, the recent Campus3D \cite{li2020campus3d} also releases large-scale photogrammetric 3D point clouds generated from high-resolution aerial images.
However, our \nicknameData{} is urban-scale and several times that of campus3d in terms of space size and labeling points.}

\subsection{3D Semantic Learning}
\bo{The wide availability of 3D datasets has facilitated rapid progress in semantic learning based on neural networks. In general, existing learning algorithms \cite{guo2019deep} can be divided into three pipelines, depending on how the 3D data is represented: \textbf{1) Voxel-based approaches} \cite{sparse, 4dMinkpwski, le2018pointgrid, Point_voxel_cnn, vvnet, zhang2020polarnet}. Although  mature 3D CNN architectures can be easily applied, these techniques usually require significant computation and memory usage, thus not being easily scalable to urban-scale point clouds. \textbf{2) 2D projection-based methods} \cite{rangenet++, lyu2020learning, salsanext, xu2020squeezesegv3}. Similarly, these pipelines leverage the well-developed 2D CNN frameworks to learn 3D semantics after projecting the point clouds onto 2D images. However, critical geometric information is very likely to be lost in the projection step, and therefore is not suitable for learning the relatively small object categories within urban-scale scenarios. \textbf{3) Point-based architectures} \cite{qi2017pointnet, qi2017pointnet++, li2018pointcnn, tangentconv, dgcnn, thomas2019kpconv, hu2019randla}.
This class of techniques learns per-point semantics primarily based on the simple MLPs and typically achieves great results in 3D object detection \cite{zhou2018voxelnet} and instance segmentation \cite{3dbonet}. Compared with both voxel and projection-based methods, these pipelines tend to be computationally efficient and have the potential to  preserve the semantics for every single 3D point. However, most of the existing point-based methods are usually designed and tuned for small-scale point sets. It is still unclear how to effectively generalize the point-based methods to the more complex urban-scale scenarios. In this regard, we investigate a number of critical challenges in Section \ref{sec:challenges}.
}

\section{The \nicknameData{} Dataset}
\label{sec:dataset}
In this section we describe how we collect, process and label the dataset over \qy{three} large urban areas in the UK.

\begin{figure}[htb]
\centering
\includegraphics[width=0.45\textwidth]{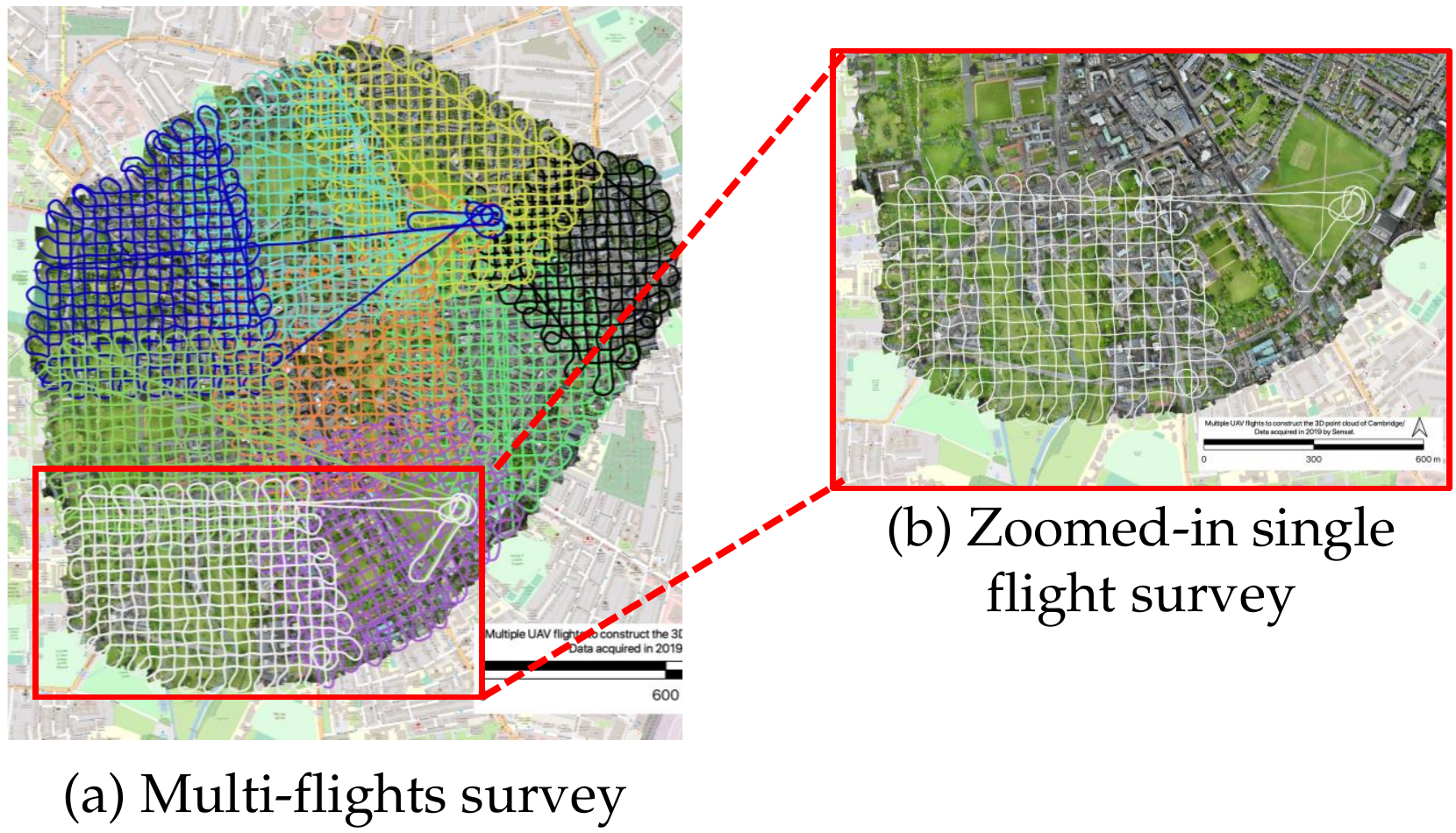}
\caption{The survey of a region in Cambridge. All 9 flight plans (\textit{left}) are collated together to cover the site. Lines with different colors represent different flight paths of UAVs. The circular path is the takeoff and landing pattern.}
\label{fig:survey}
\end{figure}

\begin{figure*}[t]
\centering
\includegraphics[width=1.0\textwidth]{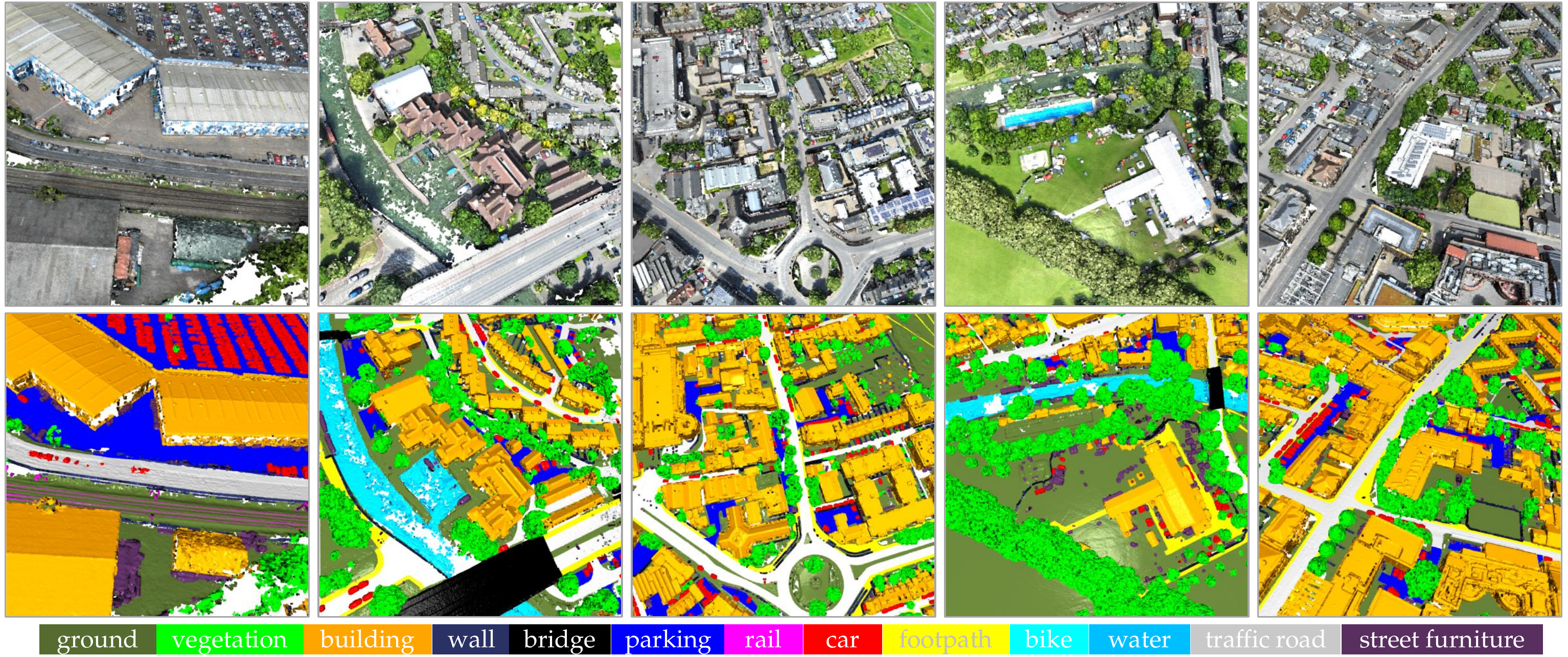}
\caption{Examples of our \nicknameData{} dataset. Different semantic classes are labeled by different colors.} 
\label{fig:annotation_examples}
\end{figure*}
\vspace{-0.1cm}

\subsection{Collecting Aerial Imagery}
\bo{Due to the clear advantages of UAV imaging over similar mapping techniques, such as LiDAR, we use a cost effective fixed wing drone, Ebee X\footnote{https://www.sensefly.com/drone/ebee-x-fixed-wing-drone/}, which is equipped with a cutting-edge SODA camera, to stably capture high-resolution aerial image sequences. In order to fully and evenly cover the survey area, all flight paths are pre-planned in a grid fashion and automated by the flight control system (e-Motion). Note that, the camera has the ability to take both oblique and nadir photographs, ensuring that vertical surfaces are captured appropriately. Since each flight lasts between 40-50 minutes due to limited battery capacity, multiple individual flights are executed in parallel to capture the whole area. These multiple aerial image sequences are then geo-referenced using a highly precise onboard Realtime Kinemtic (RTK) GNSS. Ground validation points which are measured by independent professional surveyors with high precision GNSS equipment are then used to assess the accuracy and quality of the data. For illustration, Figure \ref{fig:survey} shows the paths of the pre-planed multiple flights to cover the selected area in the Cambridge city.}

\subsection{Reconstructing 3D Point Clouds}
\bo{To reconstruct urban-scale 3D point clouds, we use off-the-shelf software such as Pix4D 
to reconstruct dense and coloured 3D point clouds from the captured aerial image sequences based on the principles of Structure from Motion (SfM) and dense image matching.}

\qy{For the urban area on the periphery of \textbf{Birmingham}, we feed all the captured sequential images to Pix4D, generating 569,147,075 3D points in total, representing an area of 1.2 square kilometers. Similarly, we reconstruct 2,278,514,725 points for the urban region adjacent to the city of \textbf{Cambridge} with an area of approximately 3.2 square kilometers, and reconstruct 904,155,619 points for \textbf{York} with an area of approximately 3.2 square kilometers.
}

\subsection{Annotating Semantic Labels}
\bo{We define the semantic categories based on two criteria. 1) Each category should have a clear and unambiguous semantic meaning, and it should be of interest to social or commercial purposes, such as asset management, urban planning, and surveillance. 2) Different categories should have significant variance in terms of geometric structure or appearance. We identify the below 13 semantic classes to label all 3D points in the Birmingham and Cambridge via off-the-shelf point cloud labeling tools. \bob{The points in York are not labelled, but made available for  possible pre-training in semi-supervised schemes.} \unclear{All labels have been manually cross-checked, guaranteeing the consistency and high quality. It takes around 600 working hours to label the entire dataset.} Figure \ref{fig:annotation_examples} shows examples of our annotations. Table \ref{tab:Overview-dataset-1} compares the statistics of our \nicknameData{} with a number of existing 3D datasets.}
\begin{enumerate}[leftmargin=*]
\itemsep -0.1cm
    \item \textit{Ground}: including impervious surfaces, grass, terrain
    \item \textit{Vegetation}: including trees, shrubs, hedges, bushes
    \item \textit{Building}: including commercial / residential buildings
    \item \textit{Wall}: including fence, highway barriers, walls
    \item \textit{Bridge}: road bridges
    \item \textit{Parking}: parking lots
    \item \textit{Rail}: railroad tracks
    \item \textit{Traffic Road}: including main streets, highways
    \item \textit{Street Furniture}: including benches, poles, lights
    \item \textit{Car}: including cars, trucks, HGVs
    \item \textit{Footpath}: including walkway, alley
    \item \textit{Bike}: bikes / bicyclists
    \item \textit{Water}: rivers / water canals
\end{enumerate}

\begin{figure}[htb]
\centering
\includegraphics[width=0.45\textwidth]{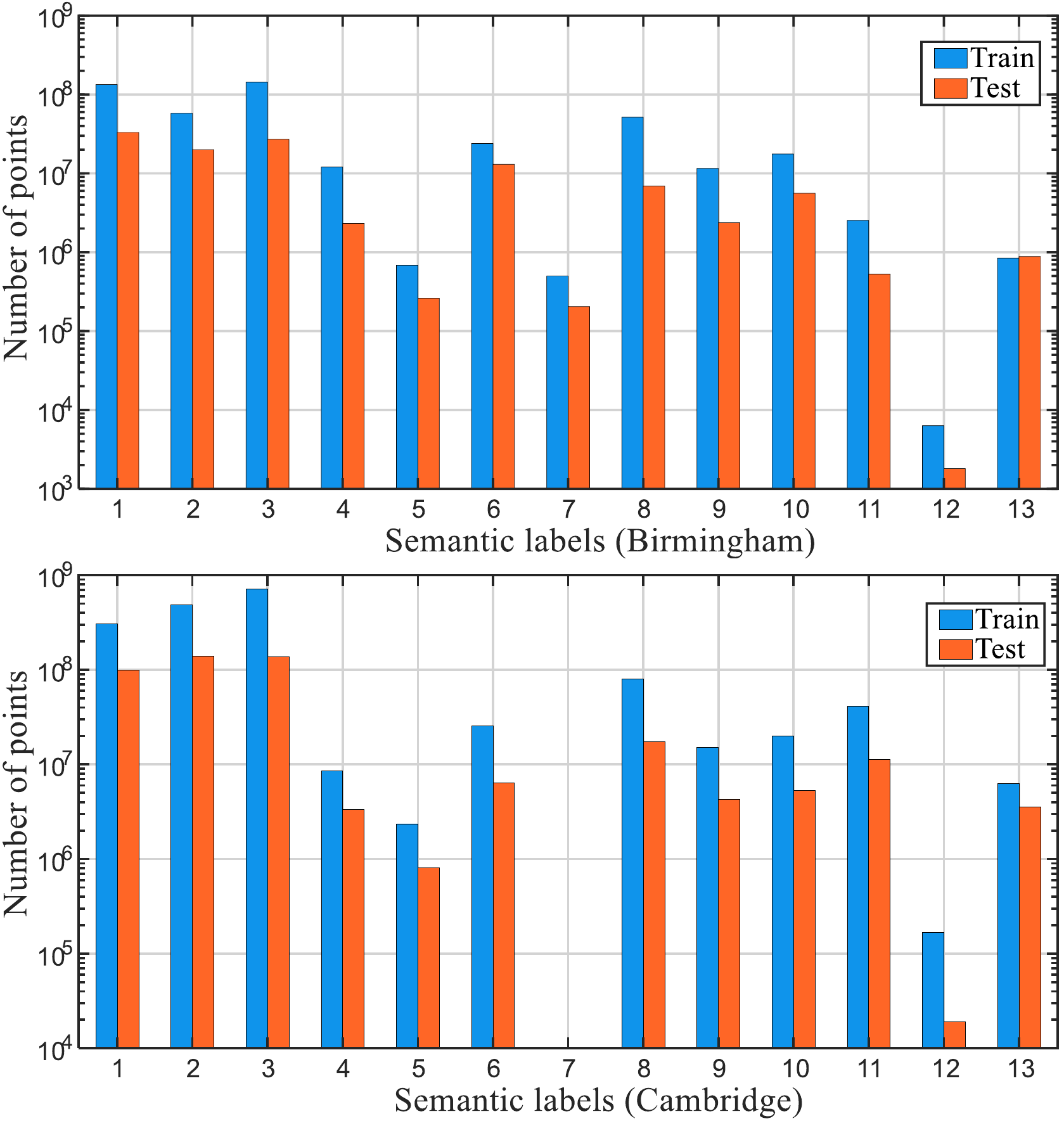}
\caption{The distribution of different semantic categories in our  \nicknameData{} dataset. Note that, there are no points annotated as \textit{rail} in Cambridge. Also note the logarithmic scale for the vertical axis.}
\label{fig:statistics}
\end{figure}

\section{Benchmarks}
\label{sec:benchmarks}
\begin{table*}[thb]
\centering
\resizebox{\textwidth}{!}{%
\begin{tabular}{r|cccccccccccccccc}
\Xhline{2.0\arrayrulewidth}
 & \rotatebox{90}{OA(\%)} &  \rotatebox{90}{mAcc(\%)} & \rotatebox{90}{\textbf{mIoU(\%)}} & \rotatebox{90}{ground} & \rotatebox{90}{veg.} & \rotatebox{90}{building} & \rotatebox{90}{wall} & \rotatebox{90}{bridge} & \rotatebox{90}{parking} & \rotatebox{90}{rail} & \rotatebox{90}{traffic.} & \rotatebox{90}{street.} & \rotatebox{90}{car} & \rotatebox{90}{footpath} & \rotatebox{90}{bike} & \rotatebox{90}{water} \\
\Xhline{1.25\arrayrulewidth}
PointNet \cite{qi2017pointnet} & 80.78 & 30.32 & 23.71 & 67.96 & 89.52 & 80.05 & 0.00 & 0.00 & 3.95 & 0.00 & 31.55 & 0.00 & 35.14 & 0.00 & 0.00 & 0.00\\
PointNet++ \cite{qi2017pointnet++} & 84.30 & 39.97 & 32.92 & 72.46 & 94.24 & 84.77 & 2.72 & 2.09 & 25.79 & 0.00 & 31.54 & 11.42 & 38.84 & 7.12 & 0.00 & 56.93\\
TagentConv \cite{tangentconv} &76.97 & 43.71 & 33.30 & 71.54 & 91.38 & 75.90 & 35.22 & 0.00 & {\ul 45.34} & 0.00 & 26.69 & 19.24 & 67.58 & 0.01 & 0.00 & 0.00 \\
SPGraph \cite{landrieu2018large} & 85.27 & 44.39 & 37.29 & 69.93 & 94.55 & 88.87 & 32.83 & 12.58 & 15.77 & \textbf{15.48} & 30.63 & 22.96 & 56.42 & 0.54 & 0.00 & 44.24 \\
SparseConv \cite{sparse} & 88.66 & 63.28 & 42.66 & 74.10 & 97.90 & {\ul 94.20} & {\ul 63.30} & 7.50 & 24.20 & 0.00 & 30.10 & {\ul 34.00} & 74.40 & 0.00 & 0.00 & 54.80 \\
KPConv \cite{thomas2019kpconv} & \textbf{93.20} & {\ul 63.76} & \textbf{57.58} & \textbf{87.10} & \textbf{98.91} & \textbf{95.33} & \textbf{74.40} & {\ul 28.69} & 41.38 & 0.00 & {\ul 55.99} & \textbf{54.43} & \textbf{85.67} & \textbf{40.39} & 0.00 & \textbf{86.30} \\
RandLA-Net \cite{hu2019randla} & {\ul 89.78} & \textbf{69.64} & {\ul 52.69} & {\ul 80.11} & {\ul 98.07} & 91.58 & 48.88 & \textbf{40.75} & \textbf{51.62} & 0.00 & \textbf{56.67} & 33.23 & {\ul 80.14} & {\ul 32.63} & 0.00 & {\ul 71.31} \\
\Xhline{2.0\arrayrulewidth}
\end{tabular}}

\caption{Benchmark results of the baselines on our \nicknameData{}. Overall Accuracy (OA, \%), mean class Accuracy (mAcc, \%), mean IoU (mIoU, \%), and per-class IoU (\%) scores are reported.\label{tab:benchmarks}}
\end{table*}


\subsection{Statistics of Train/Val/Test Split}
\bo{To setup the benchmark, we divide the point cloud of each area into similarly sized tiles similar to DALES \cite{varney2020dales}, so to be suitable for training and testing on modern GPUs. In particular, the point cloud of the Birmingham urban area is divided into 14 tiles. We then select 10 tiles for training, 2 for validation and 2 for testing. Similarly, the Cambridge split has 29 tiles in total: 20 tiles for training, 5 for validation and 4 for testing. Each tile is approximately 400$\times$400 square meters. As shown in Figure \ref{fig:statistics}, we show the total number of 3D points for each semantic category in the training/testing tiles in both Birmingham and Cambridge. It can be seen that the major semantic categories, \textit{i.e.}, \textit{ground / building / vegetation}, together comprise more than 50\% of the total points, whereas the minor yet important categories (\textit{e.g.}, \textit{bike / rail}) only account for 0.025\% of the total points. This shows that the distribution of semantic classes is extremely unbalanced, highlighting the challenges for generalizing the existing segmentation approaches.}

\subsection{Representative Baselines}
\bo{As discussed in Section \ref{sec:liter_semseg}, there are three main classes of neural pipelines to learn 3D point cloud semantics. 
In this regard, we carefully select 7 representative methods as solid baselines to benchmark our \nicknameData{} dataset.
}
\bo{
\begin{itemize}[leftmargin=*]
\itemsep -0.1cm
\item SparseConv \cite{sparse}. A solid baseline that uses submanifold sparse convolutional networks and achieves leading results on ScanNet benchmark \cite{scannet}. 
\item TagentConv \cite{tangentconv}. It projects 3D points onto tangent planes and uses 2D convolutional networks.
\item PointNet/PointNet++ \cite{qi2017pointnet,qi2017pointnet++}. These are the seminal works to directly operate on orderless point clouds. 
\item SPGraph \cite{landrieu2018large}. This is one of the first approaches capable of directly processing large-scale point clouds via the concept of superpoints.
\item KPConv \cite{thomas2019kpconv}. It introduces a flexible kernel point convolution and achieves state-of-the-art performance on the DALES dataset \cite{varney2020dales}. 
\item RandLA-Net \cite{hu2019randla}. It is the latest work for efficient semantic segmentation of large-scale point clouds and ranks the first place on Semantic3D leaderboard \cite{Semantic3D}.
\end{itemize}
}

\subsection{Evaluation Metrics}
\bo{Like the existing benchmarks \cite{Semantic3D, behley2019semantickitti, 2D-3D-S}, we use the Overall Accuracy (OA) and mean Intersection-over-Union (mIoU) as the principle evaluation metrics.}

\subsection{Benchmark Results}
\bo{For fair comparison, we faithfully follow the experimental settings of each baseline in the original publication. Table \ref{tab:benchmarks} presents the quantitative results. PointNet \cite{qi2017pointnet} has the worst performance, while KPConv \cite{thomas2019kpconv} achieves the highest mIoU scores. However, the overall segmentation performance is far from satisfactory. For example, there are still a number of key categories such as \textit{bridge, rail, street, footpath} that are poorly segmented. Furthermore, the category \textit{bike} is entirely unsegmented by all methods. Further note that different techniques have vastly different performances on these challenging categories, with no clear leader. Motivated by this, we then investigate the particular challenges that arise from our new urban-scale \nicknameData{} dataset.
}

\begin{table*}[thb]
\centering
\resizebox{\textwidth}{!}{%
\begin{tabular}{rcccccccccccccccccc}
\Xhline{2.0\arrayrulewidth}
 & Sampling & Input sets & \rotatebox{90}{OA(\%)} &  \rotatebox{90}{mAcc(\%)} & \rotatebox{90}{\textbf{mIoU(\%)}} & \rotatebox{90}{ground} & \rotatebox{90}{veg.} & \rotatebox{90}{building} & \rotatebox{90}{wall} & \rotatebox{90}{bridge} & \rotatebox{90}{parking} & \rotatebox{90}{rail} & \rotatebox{90}{traffic.} & \rotatebox{90}{street.} & \rotatebox{90}{car} & \rotatebox{90}{footpath} & \rotatebox{90}{bike} & \rotatebox{90}{water} \\
\hline
PointNet & Grid & Constant Number& \textbf{90.57} & \textbf{56.30} & \textbf{49.69} & \textbf{83.55} & \textbf{97.67} & \uline{90.66} & \textbf{22.56} & \textbf{43.54} & \uline{40.35} & \uline{9.29} & \textbf{50.74} & \textbf{29.58} & \uline{68.24} & \textbf{29.27} & 0.00 & \textbf{80.55}\tabularnewline
PointNet & Grid & Constant Volume & 88.27 & 49.80 & 42.44 & 80.20 & 96.43 & 87.88 & 8.45 & \uline{35.14} & 32.52 & 0.00 & 43.03 & 19.26 & 54.66 & 18.26 & 0.00 & 75.87\tabularnewline
PointNet & Random & Constant Number& \uline{90.34} & \uline{55.17} & \uline{48.49} & \uline{83.47} & \uline{97.51} & \textbf{90.89} & \uline{18.55} & 33.31 & \textbf{42.82} & \textbf{11.85} & \uline{47.95} & \uline{26.83} & \textbf{68.37} & \uline{29.12} & 0.00 & \uline{79.71}\tabularnewline
PointNet & Random & Constant Volume & 88.09 & 48.45 & 41.68 & 79.82 & 96.24 & 87.64 & 5.69 & 27.70 & 34.98 & 0.00 & 42.85 & 13.81 & 54.29 & 20.64 & 0.00 & 78.24\tabularnewline
\hline 
RandLA-Net & Grid & Constant Number& \textbf{91.55} & \textbf{74.87} & \textbf{58.64} & \textbf{82.99} & \textbf{98.43} & \textbf{93.41} & \textbf{57.43} & \textbf{49.47} & \textbf{55.12} & \textbf{27.33} & \textbf{60.65} & \textbf{39.43} & \textbf{84.57} & \textbf{39.48} & 0.00 & 73.97\tabularnewline
RandLA-Net & Grid & Constant Volume & 88.11 & 64.91 & 49.18 & 78.18 & 97.92 & 90.87 & 45.02 & 30.89 & 35.82 & 0.00 & 45.73 & 31.96 & 77.78 & 29.90 & 0.00 & \uline{75.30}\tabularnewline
RandLA-Net & Random & Constant Number& \uline{91.14} & \uline{74.14} & \uline{57.55} & \uline{82.25} & \uline{98.33} & \uline{92.37} & \uline{54.20} & \uline{43.10} & \uline{54.74} & \uline{25.02} & \uline{60.40} & \uline{39.17} & \uline{82.77} & \uline{37.59} & 0.00 & \textbf{78.25}\tabularnewline
RandLA-Net & Random & Constant Volume & 88.37 & 60.84 & 47.27 & 81.16 & 97.52 & 90.45 & 44.75 & 16.36 & 37.18 & 0.00 & 4219 & 26.28 & 76.76 & 30.46 & 0.00 & 71.39\tabularnewline
\Xhline{2.0\arrayrulewidth} 
\end{tabular}}
\caption{Quantitative results achieved by PointNet \cite{hu2019randla} and RandLA-Net \cite{hu2019randla} with different input preparation steps. Overall Accuracy (OA, \%), mean class Accuracy (mAcc, \%), mean IoU (mIoU, \%), and per-class IoU (\%) are reported.\label{tab:input-preparation}}
\end{table*}

\section{Challenges}\label{sec:challenges}

In this section, we identify a number of key challenges revealed by our \nicknameData{}, and explore the possible solutions to overcome them, eventually improving the segmentation performance for existing point-based approaches. Note that, we are not aiming to propose new algorithms in this section. Instead, we aim to generalize the existing pipelines from the perspective of dataset characteristics.

\subsection{Data Partition}
\bo{Due to the limited memory of modern GPUs, the first and foremost challenge is to partition the original large-scale point clouds, such that computational efficiency and segmentation accuracy can be well balanced. The early PointNet/PointNet++ techniques \cite{qi2017pointnet,qi2017pointnet++} typically divide the point clouds into 1$\times$1 meter blocks. This is however highly time consuming for such a large input tile, and causes the object geometry to be fragmented across blocks. On the other hand, if the raw point clouds are divided into extremely large blocks, the high number of points are unable to be fed into the limited GPUs. To reduce the total number of points within each block, grid or random down-sampling are applied in \cite{thomas2019kpconv, hu2019randla}. Many other methods tend to use different sampling and partitioning tricks. Overall, there is no standard and principled preparation steps in literature to partition the large-scale point clouds.}

\bo{To demonstrate the impact of different data partition schemes, we organize the data preparation into two separate steps as follows.
\begin{itemize}[leftmargin=*]
\itemsep -0.1cm
\item Step 1. To downsample the raw point clouds at the very beginning. There are two options in literature: 1) grid downsampling \cite{thomas2019kpconv}, and 2) random downsampling \cite{hu2019randla}. Both can significantly reduce the total amount of points, but each have their relative merits.
\item Step 2. To obtain individual input set of points to feed into the networks. There are two choices: 1) constant-number input sets (\textit{i.e.}, the number of points is fixed), and 2) constant-volume input sets (\textit{i.e.}, the volume of the point set is fixed). In particular, constant-number input sets are usually obtained by querying a fixed number of points with regard to the set center \cite{thomas2019kpconv,hu2019randla}, while the constant-volume input sets are extracted by collecting all points of a fixed-size cube \cite{qi2017pointnet,qi2017pointnet++}.
\end{itemize}
}
By using two representative baselines PointNet \cite{qi2017pointnet} and RandLA-Net \cite{hu2019randla}, we evaluate how the four different combinations of both Step 1 and Step 2 affect the accuracy of segmentation. 
In all the experiments, the grid size for downsampling is 0.2$m$, the random downsampling ratio is 1/10, the size for constant-volume sets is 8$\times$8$m^2$, and the constant-number sets have 4096 points.

\textbf{Analysis.} 
\bo{Table \ref{tab:input-preparation} shows the semantic segmentation scores of the eight groups of experiments on the testing split of \nicknameData{}. It can be seen that: 
\begin{itemize}[leftmargin=*]
\itemsep -0.1cm
\item Both PointNet or RandLA-Net based baselines achieve much higher scores when the input sets are number constant, compared with cases of constant volume.
\item Using grid downsampling to reduce the raw 3D point clouds demonstrates marginally better results than random downsampling for both PointNet and RandLA-Net.
\end{itemize}
}

\bo{Overall, our experiments show that the data preparation is indeed of great importance. A simple combination of grid sampling and number-consistent block partition can bring about up to 10\% improvement for mIoU scores. In this regard, we firmly believe that more studies should be conducted to further explore the effective ways for data preparation.}

\begin{table*}[thb]
\centering
\resizebox{\textwidth}{!}{%
\begin{tabular}{lcccccccccccccccc}
\Xhline{2\arrayrulewidth}
 & \rotatebox{90}{OA(\%)} &  \rotatebox{90}{mAcc(\%)} & \rotatebox{90}{\textbf{mIoU(\%)}} & \rotatebox{90}{ground} & \rotatebox{90}{veg.} & \rotatebox{90}{building} & \rotatebox{90}{wall} & \rotatebox{90}{bridge} & \rotatebox{90}{parking} & \rotatebox{90}{rail} & \rotatebox{90}{traffic.} & \rotatebox{90}{street.} & \rotatebox{90}{car} & \rotatebox{90}{footpath} & \rotatebox{90}{bike} & \rotatebox{90}{water} \\
\Xhline{2\arrayrulewidth}
PointNet \cite{qi2017pointnet} (w/o RGB) & 83.50 & 33.52 & 28.85 & 67.35 & 92.66 & 84.72 & 16.02 & 0.00 & 13.65 & 2.68 & 17.09 & 0.33 & 54.54 & 0.00 & 0.00 & 26.04\\
PointNet \cite{qi2017pointnet} (w/ RGB) & 90.57 & 56.30 & 49.69 & 83.55 & 97.67 & 90.66 & 22.56 & 43.54 & 40.35 & 9.29 & 50.74 & 29.58 & 68.24 & 29.27 & 0.00 & 80.55\\
\hdashline
PointNet++ \cite{qi2017pointnet++} (w/o RGB) & 90.85 & 56.94 & 50.71 & 79.05 & 98.37 & 94.22 & 66.76 & 39.74 & 37.51 & 0.00 & 51.53 & 38.82 & 81.71 & 5.80 & 0.00 & 65.68\\
PointNet++ \cite{qi2017pointnet++} (w RGB) & 93.10 & 64.96 & 58.13 & 86.38 & 98.76 & 94.72 & 65.91 & 50.41 & 50.53 & 0.00 & 58.40 & 46.95 & 82.31 & 38.40 & 0.00 & \textbf{82.88}\\
\hdashline
SPGraph \cite{landrieu2018large} (w/o RGB) & 84.81 & 42.12 & 35.29 & 69.60 & 94.18 & 88.15 & 34.55 & 20.53 & 15.83 & 16.34 & 31.44 & 10.54 & 55.01 & 0.98 & 0.00 & 21.57\\
SPGraph \cite{landrieu2018large} (w RGB) & 85.27 & 44.39 & 37.29 & 69.93 & 94.55 & 88.87 & 32.83 & 12.58 & 15.77 & 15.48 & 30.63 & 22.96 & 56.42 & 0.54 & 0.00 & 44.24\\
\hdashline
KPConv \cite{thomas2019kpconv} (w/o RGB) & 91.47 & 57.43 & 51.79 & 80.43 & 98.82 & 94.93 & 74.17 & 44.53 & 32.11 & 0.00 & 54.32 & 37.83 & 84.88 & 14.48 & 0.00 & 56.79\\
KPConv \cite{thomas2019kpconv} (w RGB) & \textbf{93.92} & 71.44 & \textbf{64.50} & \textbf{87.04} & \textbf{99.01} & \textbf{96.31} & \textbf{77.73} & \textbf{58.87} & 49.88 & \textbf{37.84} & \textbf{62.74} & \textbf{56.60} & \textbf{86.55} & \textbf{44.86} & 0.00 & 81.01\\
\hdashline
RandLA-Net \cite{hu2019randla} (w/o RGB) & 88.90 & 67.96 & 51.53 & 77.30 & 97.92 & 91.24 & 51.94 & 47.46 & 45.04 & 9.71 & 49.79 & 34.21 & 79.97 & 21.13 & 0.00 & 64.18\\
RandLA-Net \cite{hu2019randla} (w RGB) & 91.24 & \textbf{74.68} & 58.14 & 82.23 & 98.39 & 92.69 & 56.62 & 49.00 & \textbf{54.19} & 25.10 & 60.98 & 38.69 & 83.42 & 38.74 & 0.00 & 75.80\\
\Xhline{2\arrayrulewidth}
\end{tabular}}

\caption{Quantitative results of five selected baselines on our \nicknameData{} dataset.Overall Accuracy (OA, \%), mean class Accuracy (mAcc, \%), mean IoU (mIoU, \%), and per-class IoU (\%) are reported.\label{tab:color}}
\end{table*}

\subsection{Geometry vs. Appearance}\label{color}
\bo{One of the key differences between our \nicknameData{} and the existing LiDAR-based datasets \cite{varney2020dales, NPM3D, behley2019semantickitti} is the availability of true RGB color for every 3D point. Intuitively, the colored point clouds tend to be more informative and can provide the networks with additional features for better segmentation accuracy. However, networks may overfit the appearance and fail to learn robust features from the geometry. Taking only 3D coordinates as the input, the recent ShellNet \cite{zhang2019shellnet} achieves surprisingly good results, highlighting the power of geometry. To investigate whether and how the appearance impacts the final segmentation performance, we conduct the following ten experiments using five different baselines namely PointNet/PointNet++ \cite{qi2017pointnet,qi2017pointnet++}, SPGraph \cite{landrieu2018large}, KPConv \cite{thomas2019kpconv}, and RandLA-Net \cite{hu2019randla}}. These are either trained using only geometrical information (\textit{i.e.}, 3D coordinates) or both 3D coordinates and RGB information.

\textbf{Analysis.} 
\bo{Table \ref{tab:color} presents the quantitative results of the five baselines with respect to the different types of input point clouds. It can be seen that:
\begin{itemize}[leftmargin=*]
\itemsep -0.1cm
\item All of PointNet/PointNet++, KPConv and RandLA-Net achieve significantly better segmentation accuracy when the networks are trained given both point coordinates and RGB information. Fundamentally, this is because a number of urban classes (\textit{e.g.}, bridge, footpath, water, \textit{etc.}.) are virtually impossible to be discriminated between, if only supplied with 3D coordinates. 
\item For SPGraph, the performance depends largely on the geometrical partition which purely relies on the point coordinates, hence the inclusion of RGB does not yield a significant performance boost.
\end{itemize}
}

\bo{For all techniques, the presence of color information is critical to improve the accuracy of semantic segmentation in urban-scale scenarios. This highlights the advantage of our \nicknameData{} over the existing LiDAR based datasets such as DALES \cite{varney2020dales} and also suggests that future aerial mapping campaigns should consider including RGB. 
}

\begin{table*}[thb]
\centering
\resizebox{\textwidth}{!}{%
\begin{tabular}{lcccccccccccccccc}
\Xhline{2.0\arrayrulewidth}
 & \rotatebox{90}{OA(\%)} &  \rotatebox{90}{mAcc(\%)} & \rotatebox{90}{\textbf{mIoU(\%)}} & \rotatebox{90}{ground} & \rotatebox{90}{veg.} & \rotatebox{90}{building} & \rotatebox{90}{wall} & \rotatebox{90}{bridge} & \rotatebox{90}{parking} & \rotatebox{90}{rail} & \rotatebox{90}{traffic.} & \rotatebox{90}{street.} & \rotatebox{90}{car} & \rotatebox{90}{footpath} & \rotatebox{90}{bike} & \rotatebox{90}{water} \\
\Xhline{1.25\arrayrulewidth}
PointNet+ce &\textbf{90.57} & 56.30 & 49.69 & \textbf{83.55} & \textbf{97.67} & \textbf{90.66} & 22.56 & 43.54 & 40.35 & 9.29 & \textbf{50.74} & 29.58 & \textbf{68.24} & 29.27 & 0.00 & \textbf{80.55} \\
PointNet+wce \cite{hu2019randla} & 88.13 & \textbf{68.05} & 51.24 & 81.01 & 97.12 & 87.87 & 24.46 & {\ul 45.76} & 47.78 & \textbf{34.93} & {\ul 49.82} & 29.58 & 61.28 & 31.78 & 0.00 & 74.67 \\
PointNet+wce+sqrt \cite{salsanet} & {\ul 89.72} & {\ul 67.97} & 52.35 & {\ul 82.87} & 97.33 & {\ul 90.42} & {\ul 28.32} & 44.94 & {\ul 48.39} & 32.07 & 49.58 & \textbf{32.63} & {\ul 65.11} & {\ul 32.59} & \textbf{2.60} & 73.71 \\
PointNet+lovas \cite{berman2018lovasz} & 89.58 & 67.50 & \textbf{52.53} & 82.74 & 97.27 & 90.28 & 28.11 & 43.89 & \textbf{48.53} & {\ul 33.58} & 49.68 & 32.21 & 64.01 & \textbf{33.05} & {\ul 1.46} & {\ul 78.13} \\
PointNet+focal \cite{focal_loss} & 89.46 & 67.33 & {\ul 52.37} & 82.47 & {\ul 97.34} & 90.25 & \textbf{28.36} & \textbf{51.87} & 46.40 & 30.50 & 48.62 & {\ul 32.43} & 65.00 & 32.23 & 1.21 & 74.10 \\
\Xhline{1.25\arrayrulewidth}
RandLA-Net+ce & \textbf{93.10} & 64.30 & 57.77 & \textbf{85.39} & \textbf{98.63} & \textbf{95.40} & {\ul 62.55} & 54.85 & 56.49 & 0.00 & 58.13 & \textbf{45.90} & 82.24 & 30.68 & 0.00 & 80.70 \\
RandLA-Net+wce \cite{hu2019randla} & 91.24 & 74.68 & 58.14 & 82.23 & 98.39 & 92.69 & 56.62 & 49.00 & 54.19 & 25.10 & \textbf{60.98} & 38.69 & 83.42 & 38.74 & 0.00 & 75.80 \\
RandLA-Net+wce+sqrt \cite{salsanet} & 92.51 & \textbf{79.92} & \textbf{62.80} & 84.94 & 98.47 & {\ul 95.07} & 59.01 & \textbf{62.18} & {\ul 56.76} & {\ul 28.96} & 57.36 & 44.47 & \textbf{84.67} & {\ul 41.67} & \textbf{24.31} & {\ul 78.49} \\
RandLA-Net+lovas \cite{berman2018lovasz} & {\ul 92.56} & 76.99 & {\ul 61.51} & 84.92 & {\ul 98.55} & 94.64 & \textbf{63.17} & 52.37 & 55.43 & \textbf{36.37} & {\ul 59.35} & {\ul 45.79} & {\ul 84.28} & 41.24 & 2.66 & \textbf{80.89} \\
RandLA-Net+focal \cite{focal_loss} & 92.49 & {\ul 77.26} & 60.41 & {\ul 85.03} & 98.38 & 94.74 & 59.49 & {\ul 58.70} & \textbf{57.11} & 25.97 & 58.19 & 42.74 & 82.26 & \textbf{42.00} & {\ul 2.71} & 77.97\\
\Xhline{2.0\arrayrulewidth}
\end{tabular}}
\caption{Quantitative results achieved by PointNet \cite{qi2017pointnet} and RandLA-Net \cite{hu2019randla} with different loss functions. Overall Accuracy (OA, \%), mean class Accuracy (mAcc, \%), mean IoU (mIoU, \%), and per-class IoU (\%) are reported.\label{tab:loss}}
\end{table*}

\begin{table*}[thb]
\centering
\resizebox{\textwidth}{!}{%
\begin{tabular}{rcccccccccccccccc}
\Xhline{2.0\arrayrulewidth}
 & \rotatebox{90}{OA(\%)} &  \rotatebox{90}{mAcc(\%)} & \rotatebox{90}{\textbf{mIoU(\%)}} & \rotatebox{90}{ground} & \rotatebox{90}{veg.} & \rotatebox{90}{building} & \rotatebox{90}{wall} & \rotatebox{90}{bridge} & \rotatebox{90}{parking} & \rotatebox{90}{rail} & \rotatebox{90}{traffic.} & \rotatebox{90}{street.} & \rotatebox{90}{car} & \rotatebox{90}{footpath} & \rotatebox{90}{bike} & \rotatebox{90}{water} \\
\Xhline{1.25\arrayrulewidth}
PointNet \cite{qi2017pointnet} & 87.33 & 54.76 & 48.73 & 80.91 & 94.58 & 87.40 & 33.69 & 0.51 & 66.23 & 16.98 & 49.55 & 36.08 & 74.59 & 1.49 & 0.00 & 91.51\\
PointNet++ \cite{qi2017pointnet++} & 89.85 & 64.24 & 57.39 & 84.34 & \uline{97.11} & 89.74 & \uline{61.56} & \textbf{3.78} & 68.08 & 41.95 & 54.43 & \uline{51.54} & 84.73 & 14.43 & 0.00 & \textbf{94.34}\\
SPGraph \cite{landrieu2018large} & 80.13 & 42.87 & 36.95 & 65.75 & 93.33 & 87.24 & 41.28 & 0.00 & 42.69 & 20.94 & 2.28 & 32.05 & 64.06 & 0.00 & 0.00 & 30.76\\
KPConv \cite{thomas2019kpconv} & \textbf{91.44} & \uline{68.41} & \textbf{61.65} & \textbf{86.00} & \textbf{97.66} & \textbf{92.90} & \textbf{75.07} & 0.91 & \uline{69.74} & \textbf{55.50} & \uline{57.94} & \textbf{60.73} & \textbf{89.48} & \uline{21.44} & 0.00 & \uline{94.13}\\
RandLA-Net \cite{hu2019randla} & \uline{90.77} & \textbf{72.11} & \uline{59.72} & \uline{85.14} & 96.89 & \uline{90.77} & 59.45 & \uline{1.52} & \textbf{75.83} & \uline{48.88} & \textbf{62.58} & 48.65 & \uline{86.31} & \textbf{28.82} & 0.00 & 91.51\\
\Xhline{1.25\arrayrulewidth}
PointNet \cite{qi2017pointnet} & 86.06 & 38.56 & 29.70 & 74.94 & 94.57 & 85.38 & 8.62 & \uline{13.42} & \uline{16.47} & 0.00 & 38.64 & 14.27 & 36.96 & 0.09 & 0.00 & 2.75\\
PointNet++ \cite{qi2017pointnet++} & \uline{89.46} & 44.64 & 36.93 & \uline{77.68} & \uline{97.28} & \uline{91.95} & \uline{54.59} & 0.52 & 15.84 & 0.00 & \uline{42.08} & \uline{29.00} & \uline{67.71} & 0.24 & 0.00 & 3.16\\
SPGraph \cite{landrieu2018large} & 82.02 & 24.83 & 20.70 & 61.72 & 88.26 & 78.27 & 8.29 & 0.00 & 0.00 & 0.00 & 0.64 & 1.87 & 30.00 & 0.00 & 0.00 & 0.00\\
KPConv \cite{thomas2019kpconv} & \textbf{90.62} & \uline{48.71} & \textbf{40.51} & \textbf{78.88} & \textbf{98.33} & \textbf{94.24} & \textbf{76.20} & 0.01 & 14.70 & 0.00 & 41.77 & \textbf{39.32} & \textbf{74.22} & \uline{0.39} & 0.00 & \textbf{8.61}\\
RandLA-Net \cite{hu2019randla} & 88.92 & \textbf{51.57} & \uline{40.29} & 78.46 & 97.12 & 89.93 & 46.77 & \textbf{28.76} & \textbf{20.03} & 0.00 & \textbf{46.98} & 18.70 & 65.99 & \textbf{24.91} & 0.00 & \uline{6.15}\\
\Xhline{2.0\arrayrulewidth}
\end{tabular}}
\caption{All baselines are trained on
the Birmingham split. The top five records show the testing results on the testing split of Birmingham, while the bottom five rows show the scores on the testing split of Cambridge. Overall Accuracy (OA,
\%), mean class Accuracy (mAcc, \%), mean IoU (mIoU, \%), and per-class
IoU (\%) are reported.\label{tab:cross-city}}
\end{table*}

\subsection{The Impact of Imbalance Class Distribution}
Regardless of whether RGB is included or not, there still remain significant performance gaps between different categories. For example, the score of \textit{vegetation} is around 99\%, while the \textit{bike} is completely unable to be recognized. Fundamentally, urban areas are dominated by a small number of categories such as \textit{vegetation}, and \textit{road}, while the minor yet important classes such as \textit{bike} account for a minute portion of points. This extremely skewed distribution is another significant challenge arising from \nicknameData{}.

To alleviate this problem, a typical solution is to use more sophisticated loss functions. We evaluate the effectiveness of five off-the-shelf loss functions, with PointNet and RandLA-Net as baselines.
The loss functions are: cross-entropy, weighted cross-entropy with inverse frequency \cite{salsanext}, or with inverse square root (sqrt) frequency \cite{rosu2019latticenet}, \textit{Lov\'{a}sz}-softmax loss \cite{berman2018lovasz}, and focal loss \cite{focal_loss}.

\textbf{Analysis.} 
\bo{Table \ref{tab:loss} shows the quantitative results of the two baselines with the five different loss functions. It can be seen that the inclusion of advanced loss functions indeed improves the segmentation performance. The mIoU scores gain up to 5\%. Notably, for the extremely challenging category \textit{bike}, the baseline RandLA-Net trained with weighted cross-entropy and sqrt \cite{rosu2019latticenet} obtains more than 20\% improvement. This shows that  data imbalance can be alleviated, to an extent, by using off-the-shelf loss functions. However, even this increased performance is hardly satisfactory, and we suggest that it is still an open question to explore more effective solutions to fully tackle this challenge.}

\subsection{Cross-City Generalization}
\bo{A common issue of  deep neural networks lies in their (in)ability to directly generalize to unseen scenarios. To this end, our \nicknameData{} includes large-scale point clouds from two different urban areas, which allows us to fully evaluate their generalization ability. We conduct experiments based on 5 baselines: PointNet/PointNet++ \cite{qi2017pointnet,qi2017pointnet++}, SPGraph \cite{landrieu2018large}, KPConv \cite{thomas2019kpconv}, and RandLA-Net \cite{hu2019randla}.}
\begin{itemize}[leftmargin=*]
\itemsep -0.1cm
\item Train Birmingham/Test Birmingham: Each of the 5 baselines is only trained on the training split of Birmingham, and then tested on the testing split of the same region.
\item Train Birmingham/Test Cambridge: The above well-trained 5 baseline models are directly tested on the testing split of Cambridge. 
\end{itemize}

\textbf{Analysis.} \bo{Table \ref{tab:cross-city} compares the quantitative results of our experiments. It can be seen that the segmentation performance of all baselines drops significantly when the trained models are directly applied to novel urban scenarios. The mIoU scores have up to 20\% gaps for most approaches. Interestingly, the major categories such as ground and building do not observe severe performance drops, while the classes such as rail, street and water have the worst generalization scores. From this, we hypothesize that: 1) the imbalanced semantic distribution plays a key role in preventing the model generalization, mainly because the model tends to fit with major classes and fails to learn robust features of minor categories; 2) the more variable morphology of some urban classes such as \textit{parking} and \textit{water} are hard to be generalized from one dataset to another. Due to a lack of realistic datasets, few studies have been conducted to investigate this critical issue of generalization. It is thus an open question of how to robustly label novel urban-scale regions.}

\section{Summary and Outlook}
\label{sec:sum}
In this paper, we introduce a large and rich urban-scale \bob{dataset including two accurately labelled regions covering 4.4$km^2$ and an extra unlabelled region covering 3.2$km^2$ provided for the self/semi-supervised learning schemes.}
Through extensive benchmarking, we highlight a number of open challenges, which include how to sample and partition the large point clouds, whether to acquire RGB (color) information or not, the impact of a significantly imbalanced class distribution, and the lack of robust generalization to unseen scenarios. Other pressing challenges include instance-level and panoptic segmentation.  In the near-future, we envisage that autonomous aerial vehicles will intelligently navigate through dense cities, urban, and rural areas, and as such, real-time photogrammetric reconstruction and segmentation are also of key consideration. Accurate and high resolution 3D maps of reality are also necessary ingredients for emerging cyberphysical areas such as smart cities, intelligent transport and digital twins. It is our hope that our \nicknameData{} dataset and benchmark will be a stepping stone towards advancing research in related areas.


\noindent\textbf{Acknowledgments:} This work was partially supported by a China Scholarship Council (CSC) scholarship and the UKRI Natural Environment Research Council (NERC) Flood-PREPARED project (NE/P017134/1).

\clearpage
{
\small
\bibliographystyle{ieee_fullname}
\bibliography{egbib}
}

\clearpage
\appendix
\section*{Appendix} 
\section{Details of the Data Collection}
Our dataset is reconstructed from 2D aerial images using the well-established structure-from-motion technique, which recovers the camera extrinsic parameters for each image. The byproduct orthomosaics are only used for visualization purposes. The data are validated using GNSS RTK manual surveying carried out by professional operators. The final horizontal and vertical RMSEs are $\pm$50mm and $\pm$75mm, respectively. As a comparison, the positioning accuracy of LiDAR point clouds is around 5 to 10 cm, depending on the equipment quality, flying configuration, post-processing, etc. \cite{zhang2018patch}. We use Sensefly Soda 3D to capture the aerial images. The detailed specification of the camera can be found in Table \ref{tab:camera}. The 2D aerial images are filmed from both nadir and oblique perspectives, therefore the points on vertical surfaces are well captured. The resolution of our data depends on the number of input images and 3D reconstruction settings. Normally, photogrammetric point clouds are very dense from the process of dense image matching and so need to be subsampled. In our case, all points are subsampled at 2.5 cm, which is denser than most LiDAR data such as DALES \cite{varney2020dales}.

\begin{table}[htb]
\centering
\resizebox{0.45\textwidth}{!}{%
\begin{tabular}{r|r}
\Xhline{2\arrayrulewidth}
                                             & Specification                                   \\
\hline
Sensor size                                  & 1 inch                                       \\
RGB Lens                                     & F/2.8-11, 10.6 mm (35 mm equivalent: 29 mm)  \\
RGB Resolution                               & 5,472 x 3,648 px (3:2)                       \\
{\color[HTML]{212526} Exposure compensation} & ±2.0 (1/3 increments)                        \\
{\color[HTML]{212526} Shutter}               & Global Shutter 1/30 – 1/2000s                \\
{\color[HTML]{212526} White balance}         & Auto, sunny, cloudy, shady                   \\
{\color[HTML]{212526} ISO range}             & 125-6400                                     \\
{\color[HTML]{212526} RGB FOV}               & Total FOV: 154°, 64° optical, 90° mechanical \\
GNSS                                         & RTK/PPK                                      \\
\Xhline{2\arrayrulewidth}
\end{tabular}%
}
\caption{Detailed specifications of the camera used in our survey.}
\label{tab:camera}
\end{table}

\section{Details of the Data Annotation}
We use CloudCompare to label all the points in pure 3D. There are no unassigned points discarded in the process. To ensure the annotation quality, all annotations have been manually cross-checked. We notice that the instance annotation would be a meaningful addition to our dataset. However, due to the tremendous labeling effort of point-wise instance labels, we leave the integration of instance labels for future exploration. 

We initially labelled the point cloud as highly fine-grained 31 categories, including \textit{benches}, \textit{bollards}, \textit{road signs},  \textit{traffic lights}, \textit{etc}. Considering the scarcity of data points in certain categories, we merged some similar categories together. The initial label, merged label, and detailed mapping will be released along with the dataset.

\section{Visualization of the Dataset}
\begin{figure*}[t]
\centering
\includegraphics[width=1\textwidth]{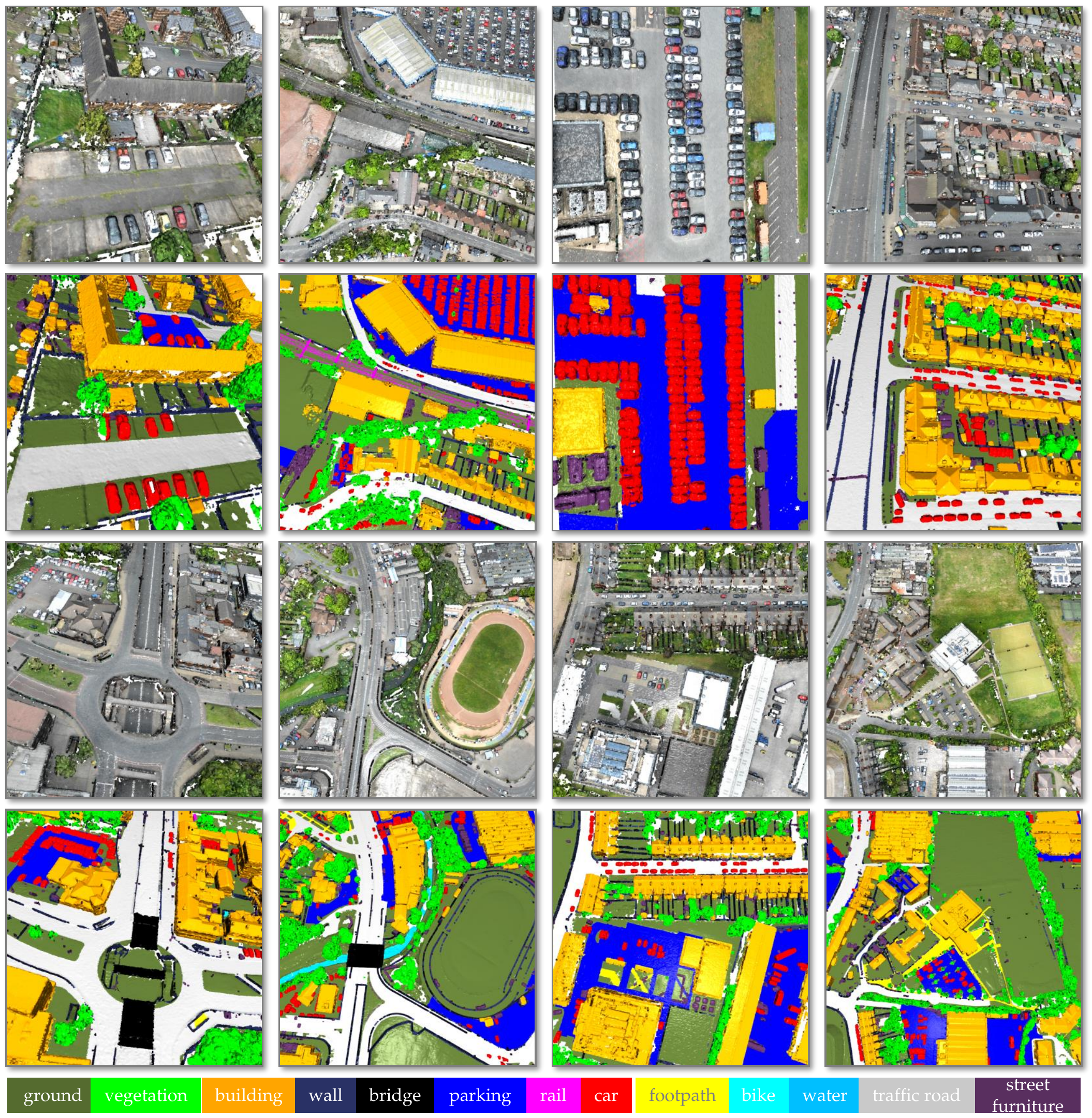}
\caption{Birmingham split of our \nicknameData{} dataset. Semantic classes are labeled by different colors.} 
\label{fig:Fig_supp_birm}
\end{figure*}

\begin{figure*}[t]
\centering
\includegraphics[width=1\textwidth]{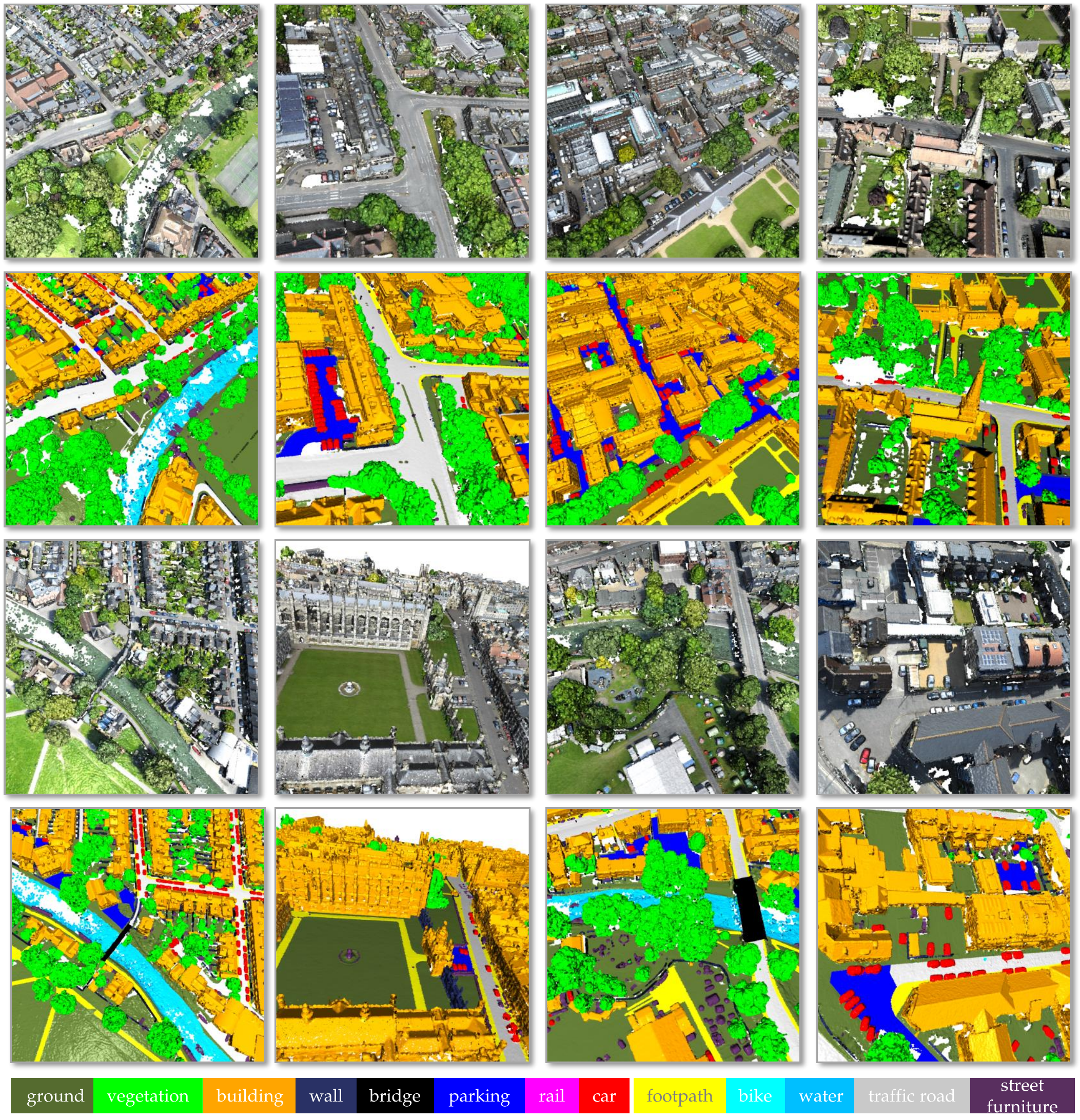}
\caption{Cambridge split of our \nicknameData{} dataset. Semantic classes are labeled by different colors.} 
\label{fig:Fig_supp_cam}
\end{figure*}

\begin{figure*}[t]
\centering
\includegraphics[width=0.8\textwidth]{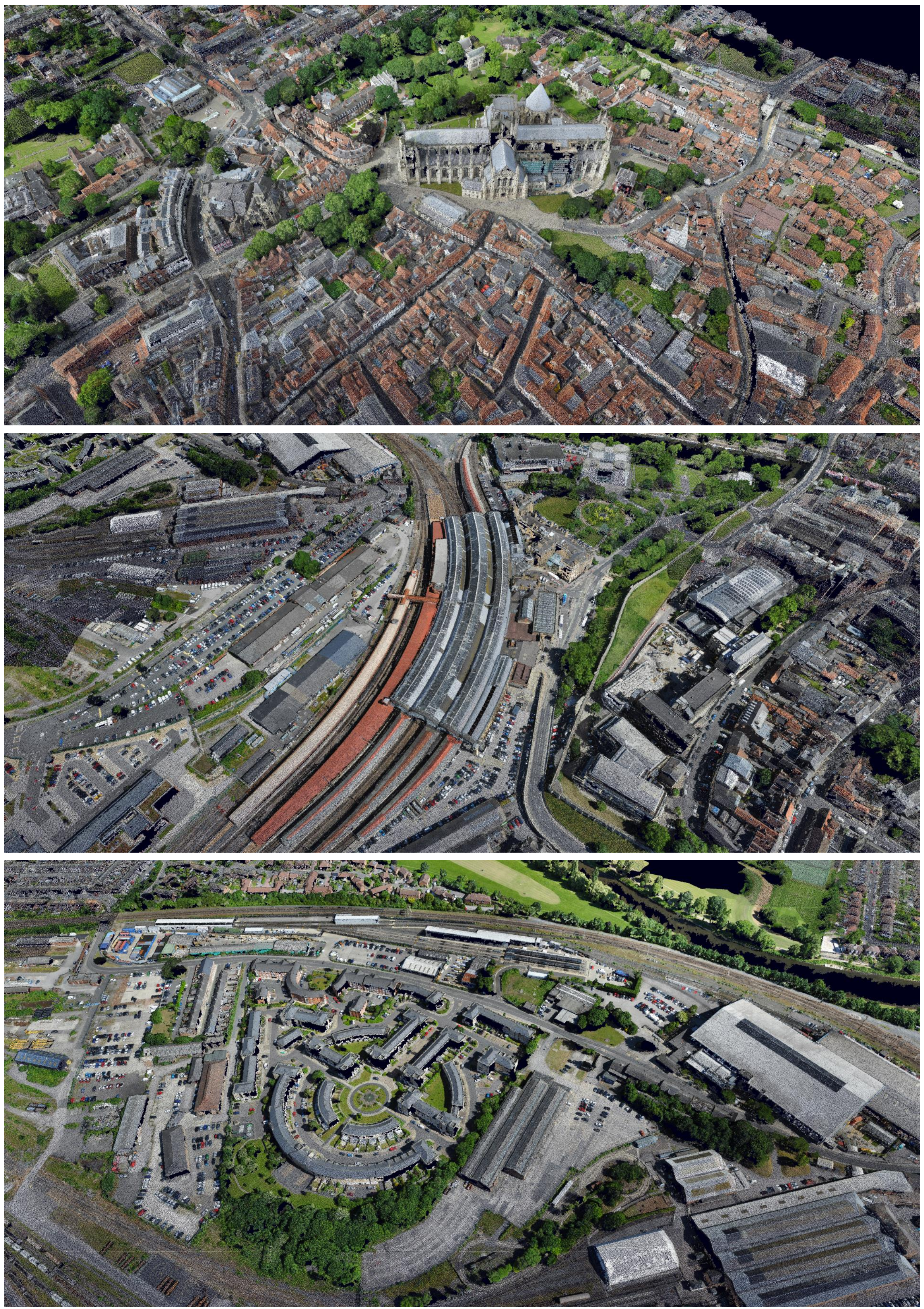}
\caption{York split of our \nicknameData{} dataset. The points in York are not labeled but made available for possible pre-training in semi-supervised or self-supervised schemes. It can be seen that our urban-scale point clouds cover various elements of a real city, such as train stations, churches, stadiums, highways, etc.} 
\label{fig:Fig_supp_York}
\end{figure*}

As mentioned in Section \ref{sec:benchmarks}, the whole urban-scale point clouds have been divided into several non-overlap tiles similar to DALES \cite{varney2020dales}. To have an intuitive and clear understanding of the data, we visualize the tiles in Birmingham and Cambridge in Figure \ref{fig:Fig_supp_birm} and Figure \ref{fig:Fig_supp_cam}, respectively. In addition, we also show some zoomed-in urban scenes from the York data in Figure \ref{fig:Fig_supp_York}.

\begin{table*}[thb]
\centering
\resizebox{\textwidth}{!}{%
\begin{tabular}{rcccccccccccccccc}
\toprule[1.0pt]
 & \rotatebox{90}{OA(\%)} &  \rotatebox{90}{mAcc(\%)} & \rotatebox{90}{\textbf{mIoU(\%)}} & \rotatebox{90}{ground} & \rotatebox{90}{veg.} & \rotatebox{90}{building} & \rotatebox{90}{wall} & \rotatebox{90}{bridge} & \rotatebox{90}{parking} & \rotatebox{90}{rail} & \rotatebox{90}{traffic.} & \rotatebox{90}{street.} & \rotatebox{90}{car} & \rotatebox{90}{footpath} & \rotatebox{90}{bike} & \rotatebox{90}{water} \\
\toprule[1.0pt]
PointNet-Rand \cite{qi2017pointnet} & 86.29 & 53.33 & 45.10 & 80.05 & 93.98 & 87.05 & \uline{23.05} & 19.52 & 41.80 & 3.38 & 43.47 & \uline{24.20} & 63.43 & 26.86 & 0.00 & \uline{79.53} \\ 
PointNet-Jigsaw \cite{jigsaw} & \uline{87.38} & \textbf{56.97} & \uline{47.90} & \uline{83.36} & \uline{94.72} & \uline{88.48} & 22.87 & \uline{30.19} & \uline{47.43} & \uline{15.62} & \uline{44.49} & 22.91 & \uline{64.14} & \textbf{30.33} & 0.00 & 77.88 \\ 
PointNet-OcCo \cite{wang2020pre} & \textbf{87.87} & \uline{56.14} & \textbf{48.50} & \textbf{83.76} & \textbf{94.81} & \textbf{89.24} & \textbf{23.29} & \textbf{33.38} & \textbf{48.04} & \textbf{15.84} & \textbf{45.38} & \textbf{24.99} & \textbf{65.00} & \uline{27.13} & 0.00 & \textbf{79.58} \\
\midrule[0.75pt]
PCN-Rand \cite{yuan2018pcn} & 86.79 & \uline{57.66} & 47.91 & \uline{82.61} & \textbf{94.82} & \uline{89.04} & \textbf{26.66} & \uline{21.96} & 34.96 & \uline{28.39} & 43.32 & \uline{27.13} & 62.97 & 30.87 & 0.00 & \uline{80.06} \\ 
PCN-Jigsaw \cite{jigsaw} & \textbf{87.32} & 57.01 & \uline{48.44} & \textbf{83.20} & \uline{94.79} & \textbf{89.25} & \uline{25.89} & 19.69 & \textbf{40.90} & \textbf{28.52} & \uline{43.46} & 24.78 & \uline{63.08} & \uline{31.74} & 0.00 & \textbf{84.42} \\ 
PCN-OcCo \cite{wang2020pre} & \uline{86.90} & \textbf{58.15} & \textbf{48.54} & 81.64 & 94.37 & 88.21 & 25.43 & \textbf{31.54} & \uline{39.39} & 22.02 & \textbf{45.47} & \textbf{27.60} & \textbf{65.33} & \textbf{32.07} & 0.00 & 77.99 \\
\midrule[0.75pt]
DGCNN-Rand  \cite{dgcnn}  & 87.54 & 60.27 & 51.96 & 83.12 & 95.43 & 89.58 & \textbf{31.84} & 35.49 & 45.11 & \uline{38.57} & 45.66 & \uline{32.97} & 64.88 & 30.48 & 0.00 & \textbf{82.34} \\ 
DGCNN-Jigsaw \cite{jigsaw} & \uline{88.65} & \uline{60.80} & \uline{53.01} & \textbf{83.95} & \textbf{95.92} & \uline{89.85} & \uline{30.05} & \textbf{43.59} & \uline{46.40} & 35.28 & \uline{49.60} & 31.46 & \uline{69.41} & \textbf{34.38} & 0.00 & \uline{80.55}\\
DGCNN-OcCo \cite{wang2020pre}   & \textbf{88.67} & \textbf{61.35} & \textbf{53.31} & \uline{83.64} & \uline{95.75} & \textbf{89.96} & 29.22 & \uline{41.47} & \textbf{46.89} & \textbf{40.64} & \textbf{49.72} & \textbf{33.57} & \textbf{70.11} & \uline{32.35} & 0.00 & 79.74 \\
\toprule[1.0pt]
\end{tabular}}
\caption{Quantitative results achieved by using OcCo \cite{wang2020pre}, Jigsaw \cite{jigsaw} and Random (Rand) initialization on the \nicknameData{} dataset, based on PointNet \cite{qi2017pointnet}, PCN \cite{yuan2018pcn} and DGCNN \cite{dgcnn} encoders. Note that, all the initialized weights are obtained by pre-training on the ModelNet40 \cite{3d_shapenets}, since these techniques are mainly designed for object-level classification and segmentation. Overall Accuracy (OA, \%), mean class Accuracy (mAcc, \%), mean IoU (mIoU, \%), and per-class IoU (\%) are reported.\label{tab:self-supervise}}
\end{table*}

\section{Additional Quantitative Results}
\subsection{Pre-training on pretext task}
Recently, a handful of works \cite{jigsaw, wang2020pre, xie2020pointcontrast} have started to design pretext tasks to achieve network pre-training based on the self-supervised learning framework. To further verify the effects of this training strategy on our urban-scale point clouds dataset, we conducted several groups of experiments on our \nicknameData{} dataset. Specifically, we evaluate the  performance of two pretraining schemes: occlusion completion \cite{wang2020pre} and context prediction \cite{jigsaw}, based on three baseline networks, including PointNet \cite{qi2017pointnet}, PCN \cite{yuan2018pcn}, and DGCNN \cite{dgcnn}. The detailed experimental results are shown in Table \ref{tab:self-supervise}.

From the results in Table we can see that, although the baseline networks are only pre-trained on the object-level point clouds, the fine-tuning model can still achieve a certain performance improvement on our dataset. In particular, the performance of several minority categories, such as \textit{rail} and \textit{bridge}, has a significant performance improvement (up to nearly 10\%), primary because the pre-trained models are less prone to overfitting to the majority categories, compared to directly training from scratch. This further demonstrates the feasibility of the pre-training strategy. However, the existing pre-training paradigm \cite{wang2020pre, jigsaw} are still limited to object-level point clouds, and it is non-trivial to be extended to large-scale point clouds. To this end, we release our unlabeled York point clouds, encouraging more studies conducted in this research area.

\section{Qualitative Results}
\begin{figure*}[thb]
\centering
\includegraphics[width=1\textwidth]{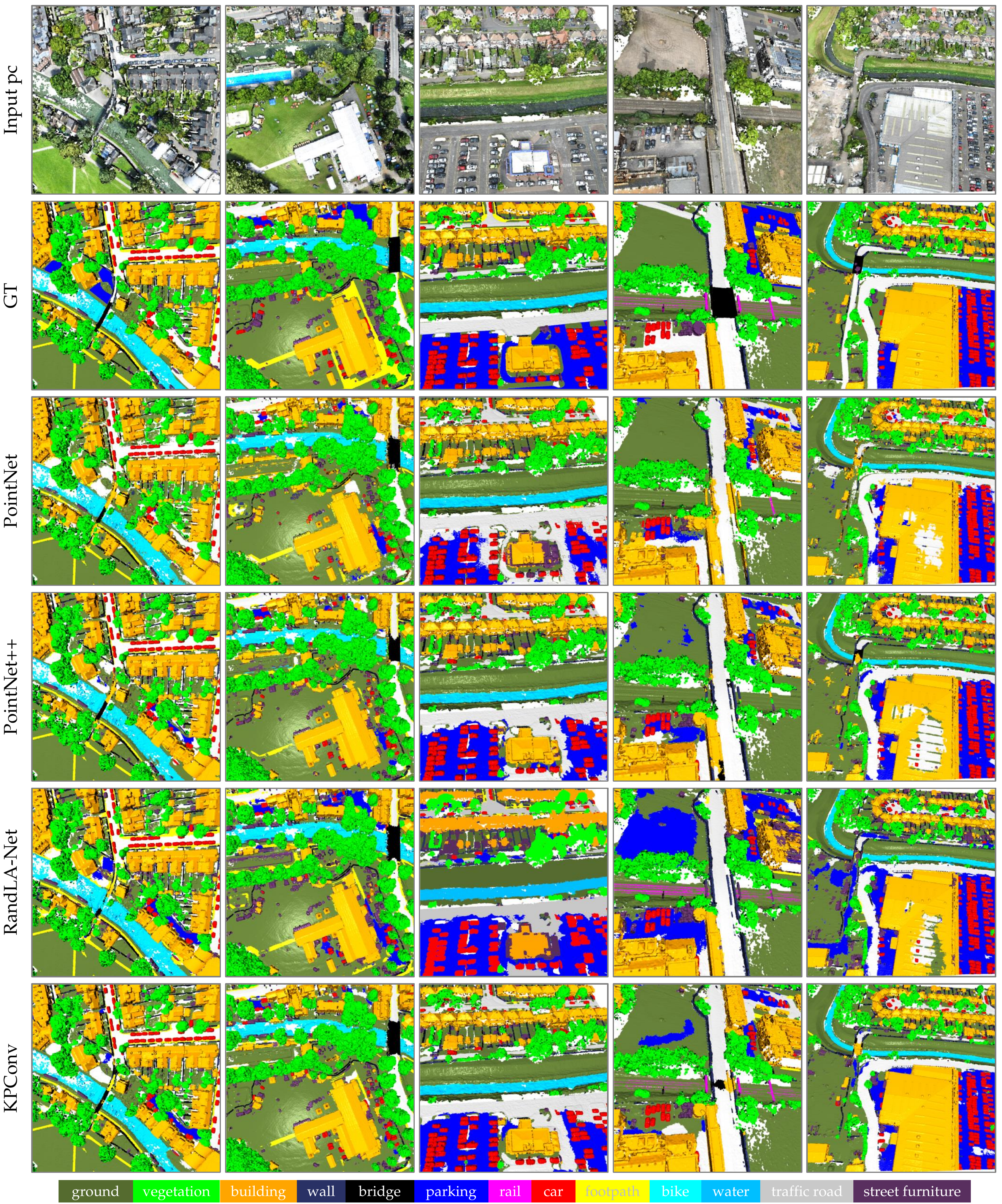}
\caption{Qualitative results of PointNet \cite{qi2017pointnet}, PointNet++ \cite{qi2017pointnet++}, RandLA-Net \cite{hu2019randla} and KPConv \cite{thomas2019kpconv}  on the test set of \nicknameData{} dataset. The black dashed box highlights the inconsistency predictions with the ground-truth label.}
\label{fig:Fig_supp_results}
\end{figure*}

We also show the corresponding qualitative results achieved by several baselines on the test set of our \nicknameData{} in Figure \ref{fig:Fig_supp_results}. The detailed quantitative results can be found in Section \ref{color}.

\section{Video Illustration} 
We provide a video demo illustrating our \nicknameData{} dataset, which can be viewed at \url{https://www.youtube.com/watch?v=IG0tTdqB3L8&t=5s}.

\end{document}